\pdfoutput=1
\documentclass[11pt]{article}

\usepackage[preprint]{acl} 

\usepackage{times}
\usepackage{latexsym}

\usepackage[T1]{fontenc}

\usepackage[utf8]{inputenc}

\usepackage{microtype}

\usepackage{inconsolata}

\usepackage{graphicx}

\usepackage{linguex}
\usepackage{array}
\usepackage{enumitem}
\usepackage{colortbl}
\usepackage{booktabs}
\usepackage{amsmath}
\usepackage{float}
\usepackage{fontawesome}
\usepackage{CJKutf8}
\usepackage{placeins}

\newcommand{\faCheckGreen}{{\color{green}\faCheck}}
\newcommand{\faTimesRed}{{\color{red}\faTimes}}

\definecolor{good}{rgb}{0.0, 1.0, 0.0}
\definecolor{bad}{rgb}{1.0, 0.0, 0.0}

\definecolor{mygreen}{rgb}{0,0.6,0}
\definecolor{mygray}{rgb}{0.5,0.5,0.5}
\definecolor{mymauve}{rgb}{0.58,0,0.82}

\title{Inconsistent Tokenizations Cause Language Models to be Perplexed by Japanese Grammar}

\author{
 \textbf{Andrew Gambardella},
 \textbf{Takeshi Kojima},
 \textbf{Yusuke Iwasawa},
 \textbf{Yutaka Matsuo}
\\
 University of Tokyo
\\
 \small{
   \textbf{Correspondence:} \href{mailto:atgambardella@weblab.t.u-tokyo.ac.jp}{atgambardella@weblab.t.u-tokyo.ac.jp}
 }
}

\begin{document}
\maketitle
\begin{abstract}
Typical methods for evaluating the performance of language models evaluate their ability to answer questions accurately. These evaluation metrics are acceptable for determining the extent to which language models can understand and reason about text in a general sense, but fail to capture nuanced capabilities, such as the ability of language models to recognize and obey rare grammar points, particularly in languages other than English. We measure the perplexity of language models when confronted with the ``first person psych predicate restriction'' grammar point in Japanese. Weblab is the only tested open source model in the 7-10B parameter range which consistently assigns higher perplexity to ungrammatical psych predicate sentences than grammatical ones. We give evidence that Weblab’s uniformly bad tokenization is a possible root cause for its good performance, and show that Llama 3’s perplexity on grammatical psych predicate sentences can be reduced by orders of magnitude (28x difference) by restricting test sentences to those with uniformly well-behaved tokenizations. We show in further experiments on machine translation tasks that language models will use alternative grammar patterns in order to produce grammatical sentences when tokenization issues prevent the most natural sentence from being output.
\end{abstract}

\section{Introduction}

It is unusually difficult to find the correct tests and metrics to evaluate large language model (LLM) performance. Standard benchmarks such as MMLU~\cite{Hendrycks2021MMLU} are designed to capture the memorization and reasoning abilities of LLMs, but as LLMs are general tools which can be used to solve a large variety of tasks, there are many aspects of LLM performance which these benchmarks cannot cover.
\begin{figure}[h]
    \centering
    \includegraphics[width=\linewidth]{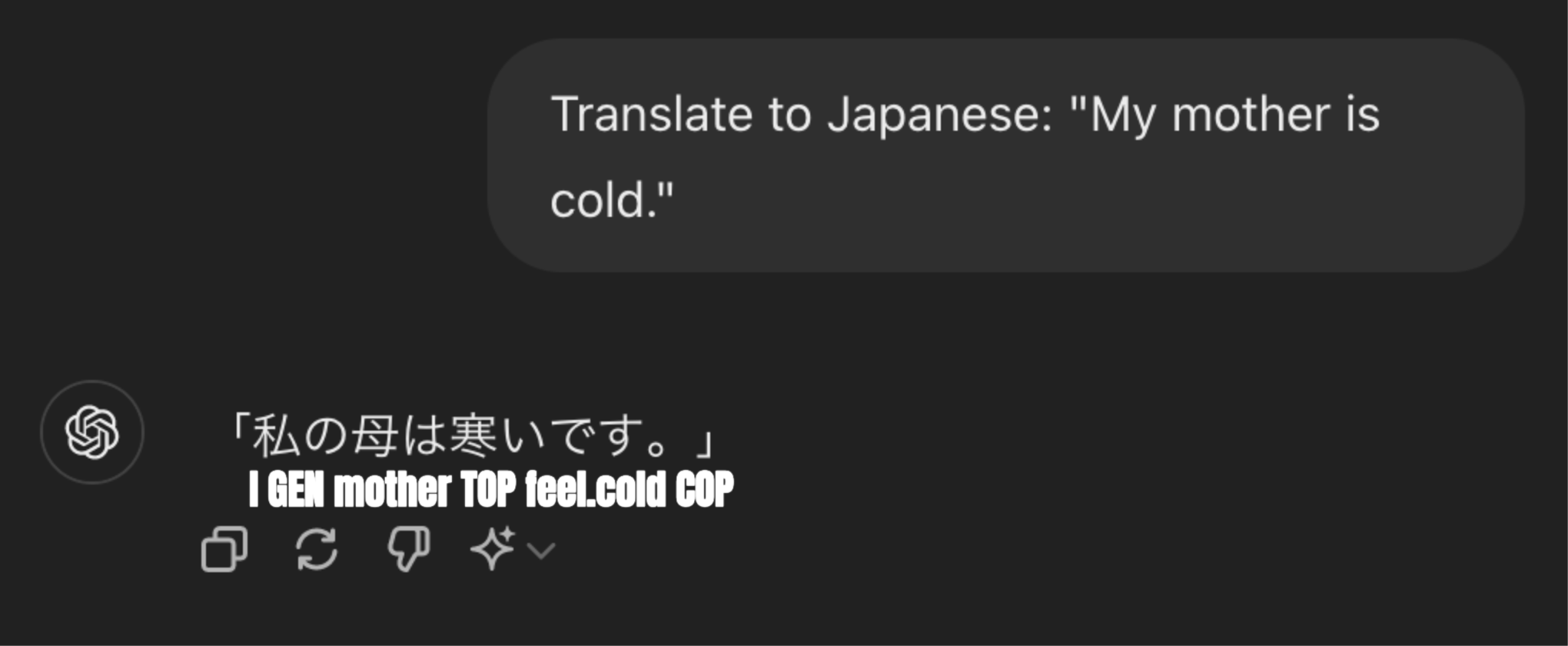}
    \caption{State of the art language models frequently fail to respect nuanced aspects of Japanese grammar, such as the first person psych predicate restriction, where here GPT-4o produces a sentence which is functionally identical to the ungrammatical Example~\ref{ex:ungrammaticalpsych} in Section~\ref{sec:psychrestriction}.}
    \label{fig:gpt4psychpredicate}
\end{figure}

In particular, these benchmarks are generally English-centric, and tend to test ``higher-level'' abilities, whereas more nuanced evaluation of LLMs has received much less attention. Even for other languages such as Japanese, existing benchmarks such as llm-jp-eval~\cite{han2024llmjpeval} tend to test these higher-level abilities in Japanese, and do not test any Japanese-specific abilities. This is concerning, because Japanese contains several rare grammatical structures that require special attention, especially considering that many foundation models are now trained on multilingual datasets, and one would assume some degree of cross-lingual information sharing due to the prevalence of linguistic universals~\cite{GreenbergJH1966}. We inspect the perplexities of pretrained LLMs in order to study nuanced grammatical corner cases in Japanese, which are frequently studied in linguistics literature, but for which it would be difficult to create a large dataset. In particular, we study the ``first person psych predicate restriction'' with models in the 7-10B parameter range. We find that of open source models that currently exist, only Weblab has lower perplexities on grammatical sentences compared to ungrammatical sentences containing psych predicates. We provide evidence that this is due to Weblab's uniformly bad tokenizations, and show that Llama 3 performs orders of magnitude better on this task when its test set is restricted to sentences which it tokenizes uniformly well. We further show evidence that language models will, when tasked to output sentences containing third-person psych predicate expressions, utilize unusual circumlocutions or grammar patterns. These findings suggest that more consistent tokenizers for Japanese could lead to LLMs which are able to more closely obey relatively obscure Japanese grammar rules and output sentences utilizing more natural grammatical constructions.

\section{Linguistic Preliminaries}

\subsection{First Person Psych Predicate Restriction}
\label{sec:psychrestriction}

Japanese is notable for having what is referred to as a ``first person psych predicate restriction,'' meaning that certain predicates which describe the internal states of people (such as ``happy,'' ``sad,'' ``cold,'' ``dizzy,'' etc.) can only be used to describe the first person in general when used directly.
When describing others, evidential expressions (such as ``seems,'' ``appears,'' etc) must be used~\cite{hasegawa2005self,lee2012psych}. \citet{hasegawa2005self} give the following examples:

\exg. Watashi wa samu-i. \label{ex:firstpersonspsych}\\
I TOP feel.cold-NPST  \\
\glt ``I feel cold.''

\exg. \#Haha wa samu-i. \label{ex:ungrammaticalpsych}\\
Mother TOP feel.cold-NPST  \\
\glt ``My mother feels cold.'' [Intended]

\exg. Haha wa samu-gat-te i-ru.  \label{ex:thirdpersonpstchevid1}\\
Mother TOP feel.cold-EVID-CONJ be-NPST\\
\glt Lit. ``My mother is showing signs of feeling cold.''

\exg. Haha wa samu-soo da. \label{ex:thirdpersonpstchevid2}\\
Mother TOP feel.cold-EVID COP.NPST  \\
\glt ``My mother appears to feel cold.''

Native speakers of Japanese generally obey the first-person psych predicate restriction in all cases, even if they are not consciously aware of its existence. L2 speakers of Japanese, however, are frequently unaware of it, which can lead them to produce utterances that are extremely confusing to native speakers. Similar grammatical phenomena and L2 misuse have been observed in Korean, which features a comparable first-person psych predicate restriction~\cite{ahn2024psych}.

\begin{table*}[!htbp]
\centering
\begin{tabular}{lcccccc}
\toprule
    & \textbf{Mistral} & \textbf{Llama 2} & \textbf{Llama 3} &  \textbf{Weblab} & \textbf{Swallow} & \textbf{Swallow-MS} \\
\midrule
(\#) 3rd, psych, direct & 2.0e+04 & 3.3e+04 & 6.9e+03 & 2.0e+06 & 1.2e+03 & 1.9e+03 \\\midrule
(a) 1st, psych, direct & \cellcolor{bad!20}3.6e+04 & \cellcolor{bad!20}1.2e+05 & \cellcolor{bad!20}9.1e+04 & \cellcolor{good!20}6.1e+05 & \cellcolor{bad!20}1.9e+03 & \cellcolor{bad!20}3.2e+03 \\
(b) 3rd, non-psych, direct & \cellcolor{green!20}1.8e+03 & \cellcolor{good!20}5.9e+03 & \cellcolor{good!20}4.5e+03 & \cellcolor{good!20}7.3e+05 & \cellcolor{yellow!20}1.2e+03 & \cellcolor{bad!20}2.9e+03 \\
(c) 3rd, psych, evidential & \cellcolor{yellow!20}2.0e+04 & \cellcolor{bad!20}4.9e+04 & \cellcolor{bad!20}3.7e+04 & \cellcolor{green!20}1.3e+06 & \cellcolor{bad!20}4.1e+03 & \cellcolor{bad!20}3.3e+03 \\
\bottomrule
\end{tabular}
\caption{Median perplexity over language models, for sentences corresponding to those introduced in Section~\ref{sec:psychrestriction}. Weblab is the only model which has lower perplexities for all grammatical constructions (labeled a, b, c) relative to the ungrammatical direct third person psych predicate (labeled \#), which we believe is due to its uniformly bad tokenization. Green, yellow, and red indicate perplexity for grammatical constructions that are respectively lower than, equal to, and higher than that of the grammatical constructions.}
\label{tab:psych_predicates}
\end{table*}

As \citet{hasegawa2005self} further note, when the predicate is polysemous, the semantic role of the subject shifts to conform to this restriction (so that Example~\ref{ex:ungrammaticalpsych} must be interpreted as ``I, the mother, am cold'' or possibly ``I am cold because of my mother''). There are also exceptions for the nonreportive style (such as in a novel, where the narrator can know the internal states of all characters) where third-person subjects are permitted for psych predicate expressions~\cite{kuroda1973where}. From the perspective of LLM training, checking whether or not a model obeys the first person psych predicate restriction is interesting, as we must test whether the LLM has learned to \emph{not} do something which was \emph{not} in the training dataset, rather than to reproduce co-occurrences and patterns which \emph{did} exist in the dataset. Notably, even state of the art models such as GPT-4o~\cite{OpenAIGPT42023} frequently create ungrammatical or strange sentences which fail to respect this grammatical rule, as shown in Figure~\ref{fig:gpt4psychpredicate}.

\section{Experiments}
\label{sec:experiments}

We wish to see whether language models have the capability to understand the grammar rules introduced in Sections~\ref{sec:psychrestriction}. For this case, it is simple to create minimal pairs of sentences similar to Examples~\ref{ex:firstpersonspsych},  \ref{ex:thirdpersonpstchevid1}, and \ref{ex:thirdpersonpstchevid2}, which can be compared against ungrammatical constructions similar to Example~\ref{ex:ungrammaticalpsych}. We evaluate the Huggingface~\cite{wolf2019huggingface} implementations of the following LLMs in the 7B to 10B parameter range: Mistral 0.1-7B~\cite{jiang2023mistral}, Llama 2-7B~\cite{touvron2023llama}, Llama 3-8B~\cite{llamateam2024llama3}, Weblab-10B\footnote{https://huggingface.co/matsuo-lab/weblab-10b}, Swallow-7B~\cite{fujii2024swallow}, and Swallow-MS-7b-v0.1\footnote{https://huggingface.co/tokyotech-llm/Swallow-MS-7b-v0.1}. Mistral, Llama 2, and Llama 3 are multilingual LLMs, whereas Weblab, Swallow, and Swallow-MS are specifically tuned for Japanese applications. In Section~\ref{sec:psychrestrictionperplexityexps} we choose to use perplexity on base models as our measure because it can be easily applied on the sentence level regardless of how the sentence is tokenized, and also correlates directly with token generation probabilities (e.g., sentence x has lower perplexity than y generally implies that sentence x is more likely to be generated than y). We also report the median over perplexity scores rather than the mean as the scales and numerical values of token probabilities can differ greatly between models due to variations in training data, vocabulary sizes, and tokenizer designs (including differing fertility scores, as discussed below). These differences tend to affect the mean much more significantly due to its sensitivity to outliers, making the median a more robust statistic for comparing central tendencies across diverse models in this context. In Section~\ref{sec:psychrestrictionmtlexps} we further evaluate the instruction-following variants of these models when presented with machine translation tasks.

\begin{table*}[!htbp]
\centering
\begin{tabular}{lcccccc}
\toprule
    & \textbf{Mistral} & \textbf{Llama 2} & \textbf{Llama 3} &  \textbf{Weblab} & \textbf{Swallow} & \textbf{Swallow-MS} \\
\midrule
Fertility & 1.51 & 1.58 & 0.85 & 1.23 & 1.0 & 1.0 \\\midrule
Byte Fallback & 0.43 & 0.49 & 0.08 & 0.66 & 0.19 & 0.19 \\
\bottomrule
\end{tabular}
\caption{Japanese fertility scores and byte fallback rates across studied models over sentences produced by the templates given in Appendix~\ref{sec:psychappendix}. Due to its use of an unmodified English tokenizer, the majority of tokens produced by Weblab are out-of-vocabulary.}
\label{tab:fertility_scores}
\end{table*}

\subsection{Perplexity Experiments}
\label{sec:psychrestrictionperplexityexps}

For the psych predicate restriction, an ideal language model would have the following properties:

\begin{enumerate}[label=(\alph*)]
    \item The probability of an adjective or verb which is not a psych predicate given that the sentence is in the third person is higher than the probability of an adjective or verb which is a psych predicate given that the sentence is in the third person.
    \item The probability of an adjective or verb which is a psych predicate given that the sentence is in the first person is higher than the probability of the same adjective or verb given that the sentence is in the third person.
    \item The probability that a psych predicate is used with an evidential given that the sentence is in the third person is higher than the probability of the same psych predicate being used directly given that the sentence is in the third person.
\end{enumerate}

We show perplexities corresponding to the grammatical constructions in (a), (b), and (c) above, as well as the ungrammatical third person direct psych predicate expression, in Table~\ref{tab:psych_predicates} with templates for our test sentences in Appendix~\ref{sec:psychappendix}. Despite having much higher perplexities in general, only Weblab has lower perplexities for the grammatical constructions than the ungrammatical construction. We found that our results were influenced heavily by tokenization, particularly whether or not byte fallback (which reduces the probabilities of tokens by several orders of magnitude) was needed, and whether or not the tokenization corresponded to the underlying linguistic structure. In particular, we noticed that for Llama 3, while certain adjectives (``itai,'' ``samui,'' ``tsurai,'' ``kokorobosoi,'' ``atsui'') caused well-behaved tokenizations, those ending in ``shii'' such as (``kanashii,'' ``sabishii,'' ``ureshii,'' ``hazukashii,'' ``kurushii'') caused byte fallback when used in conjunction with evidential expressions, thus giving these tokens extremely low probability, even when they are grammatically necessary. When restricting our test sentences to only those which Llama 3 tokenizes well, we found that its median perplexity for evidentials used with psych predicates in the third person (sentence type (c)) drops from the reported 3.7e+04 to 1.3e+03 (\textasciitilde28x difference), whereas the perplexity of the ungrammatical construction drops from the reported 6.9e+03 to only 3.9e+03 (\textasciitilde1.8x difference). Consequently, the Llama 3 base model is very likely to output the ungrammatical direct form of a psych predicate adjective with a third person topic, if and only if that adjective ends in ``shii'' as opposed to just ``i.'' Note that for sentence type (c) compared to the ungrammatical construction only the final tokens in the sentence differ, so lower perplexity is exactly equivalent to higher generation chance.

\begin{table*}[!htbp]
\centering
\begin{tabular}{lcccc|cccc}
\toprule
    & \multicolumn{4}{c|}{\textbf{Weblab}} & \multicolumn{4}{c}{\textbf{Llama 3}} \\
    \cmidrule(r){2-5} \cmidrule(l){6-9}
    & \textbf{Cold} & \textbf{Embarrassed} & \textbf{Lonely} & \textbf{Pain} & \textbf{Cold} & \textbf{Embarrassed} & \textbf{Lonely} & \textbf{Pain} \\
\midrule
\faCheckGreen evidential & 47 & 90 & 0 & 0 & 0 & 32 & 0 & 0 \\
\faCheckGreen grammatical & 53 & 10 & 0 & 6 & 0 & 0 & 0 & 0 \\
\faTimesRed no evidential & 0 & 0 & 100 & 94 & 69 & 39 & 100 & 100 \\
\faTimesRed mistranslation & 0 & 0 & 0 & 0 & 0 & 29 & 0 & 0 \\
\faTimesRed wrong syntax & 0 & 0 & 0 & 0 & 31 & 0 & 0 & 0 \\
\bottomrule
\end{tabular}
\caption{Weblab and Llama 3 outputs when asked to translate the English sentence ``My mother is \{psych predicate\}'' into Japanese. While Llama 3 struggled to output evidential expressions at all, Weblab was able to consistently output evidential expressions with a third person subject feeling ``cold'' or ``embarrassed.'' Here ``grammatical'' indicates alternative phrasings that are grammatical translations of the sentence, but do not require the use of evidential expressions.}
\label{tab:mtl_exp}
\end{table*}

Ironically, Weblab may have performed best on these tasks because it was trained using a tokenizer for English models as-is, and as such does not contain tokens for even extremely common Japanese characters\footnote{Weblab's tokenizer uses byte fallback for words such as ``taberu'' (to eat) and ``kau'' (to buy), the characters for which are learned in the second year of elementary school.}. As a result, it tokenizes every sentence uniformly poorly, and does not have grammar-specific tokenization issues. To quantitatively support claims about tokenizer strength, we report fertility scores~\cite{rust2021tokenizer} and byte fallback rates in Table~\ref{tab:fertility_scores}. Weblab, which used an unmodified English language tokenizer, must output significantly more tokens, and outputs out-of-vocabulary tokens much more frequenly than LLMs such as Llama 3, which use tokenizers more adapted to Japanese.
These findings suggest that the same issues which seem to appear in very large state of the art models, such as GPT-4o given in Figure~\ref{fig:gpt4psychpredicate}, could possibly be alleviated simply by using more consistent tokenizers or, alternatively, using uniformly bad tokenizers (such as only bytes).

\subsection{Machine Translation Experiments}
\label{sec:psychrestrictionmtlexps}

We further conducted machine translation experiments with the instruction-following variants of Weblab and Llama 3, on the sentence ``My mother is \{psych predicate\}'' with the four psych predicates: cold, embarrassed, lonely, and pain. In these experiments, we collect 100 samples using a temperature of $0.4$ to encourage a distribution of outputs which reflect the most likely outputs of each model. The authors labeled each output of the model, categorizing them into one of \{evidential, grammatical (non-evidential), no evidential (ungrammatical), mistranslation, wrong syntax\}, and summarized these results in Table~\ref{tab:mtl_exp}.

As the perplexity experiments in Section~\ref{sec:psychrestrictionperplexityexps} suggested, while Weblab was able to output evidential expressions, or otherwise grammatical expressions, fairly often, Llama 3 very rarely used evidential expressions, and failed to output grammatical expressions in most cases. Furthermore, unlike Weblab, Llama 3 also occasionally produced sentences which were semantically incorrect given the English sentence (mistranslations) or syntactically incorrect (invalid Japanese), despite Llama 3 generally outperforming Weblab by a large margin in Japanese tasks. Notably, the instruction-following Llama 3 output ungrammatical expressions even in the case in which the base model had lower perplexity for grammatical expressions relative to ungrammatical expressions (the psych predicate adjectives which do not end in ``shii'' such as ``itai'' (feels pain) and ``samui'' (feels cold)). In the machine translation setup, both the weights of the model and prompts used differ drastically from those in the perplexity experiments, which makes this difference unsurprising. Nevertheless, the general trend that Weblab was able to output grammatical expressions featuring evidentials when psych predicates were used in the third person, despite its weaker tokenizer and even in this vastly different scenario, points to our claim that inconsistent tokenizations are the root cause behind LLMs making this grammatical error in Japanese.

\section{Conclusion}

We constructed a set of minimal pair test sentences to measure the Japanese grammar abilities of several open source models, particularly in their ability to recognize the ``first person psych predicate restriction.'' For these minimal pairs, we showed that inconsistent tokenizations cause language models to produce perplexities that do not match the grammatical rules. Only when given uniformly good, or uniformly bad, tokenizations of Japanese were these models able to produce lower perplexity for grammatical psych predicate constructions relative to ungrammatical psych predicate constructions. Our findings should inform future constructions of tokenizers and pretraining datasets, which ultimately should lead to future language models being able to more closely follow nuanced grammar rules in Japanese and other languages.

\section{Limitations}

While we have done our best to come up with explanations for the phenomena which we have observed, the language models studied have many confounding factors and unknowns which could be altering token probabilities in ways that we could not know. For instance, in general we cannot not know the exact effects of the amount of Japanese pretraining data, the ratio of Japanese pretraining data, and the effect of tokenizations on the produced token probabilities. While the existence of byte fallback played a crucial role in our analysis, it also serves as a confounding factor, as it causes single Japanese characters to be split into multiple tokens, and makes the probabilities for those tokens extremely low.

\bibliography{references}

\newpage

\appendix

\section{Interlinear Notation Key}

\begin{tabular}{>{\raggedright}p{1.5cm} p{5cm}}
ACC & accusative \\
COP & copula \\
CONJ & conjunctive \\
EVID & evidential \\
GEN & genitive \\
LAT & lative \\
NMLZ & nominalizer \\
NPST & nonpast tense \\
NOM & nominal \\
PST & past \\
TOP & topic marker \\
\end{tabular}

\section{Section~\ref{sec:psychrestrictionperplexityexps} Psych Predicate Experiment Template}
\label{sec:psychappendix}

The template used to generate sentences for the psych predicate experiments is as follows:

\exg. \{Watashi/Haha\} wa \{predicate\}-\{i/soo\}. \\
\{I/Mother\} TOP \{predicate\}-\{NPST/EVID\}  \\
\glt ``\{I/My mother\} \{feel(s)/appears to feel\} \{predicate\}.''

\subsection{Python Code for Dataset Generation}
The Python code in Figure~\ref{fig:pythoncode} can be used to generate the example sentences based on the template structure.

\begin{figure*}[htbp] 
\centering 
\begin{CJK}{UTF8}{min} 
\footnotesize 
\begin{verbatim}
def generate_sentences() -> None:
    subjects = {
        "watashi": "私",  # I
        "haha": "母"     # Mother
    }

    # Psych predicates and non-psych predicates
    predicates = {
        "yasashii": "優しい",      # nice (non-psych)
        "kawaii": "可愛い",        # cute (non-psych)
        "osoi": "遅い",            # late/slow (non-psych)
        "hayai": "速い",           # fast (non-psych)
        "chikai": "近い",          # close (non-psych)
        "wakai": "若い",           # young (non-psych)
        "omoshiroi": "面白い",     # interesting (non-psych)
        "segatakai": "背が高い",  # tall (non-psych)
        "urusai": "うるさい",      # loud/annoying (non-psych)
        "tsuyoi": "強い",          # strong (non-psych)
        "itai": "痛い",            # feel.pain (psych)
        "samui": "寒い",           # feel.cold (psych)
        "tsurai": "辛い",          # feel.difficult/painful (psych)
        "kokorobosoi": "心細い",  # feel.helpless (psych)
        "atsui": "暑い",           # feel.hot (psych)
        "kanashii": "悲しい",      # feel.sad (psych)
        "sabishii": "寂しい",      # feel.lonely (psych)
        "ureshii": "嬉しい",       # feel.happy (psych)
        "hazukashii": "恥ずかしい", # feel.embarrassed (psych)
        "kurushii": "苦しい"       # feel.strenuous/painful (psych)
    }

    for subj_key, subject_jp in subjects.items():
        for pred_key, predicate_jp in predicates.items():
            # Generate direct form (-i)
            # Grammatical for 1st person (watashi) + any predicate
            # Grammatical for 3rd person (haha) + non-psych predicate
            # Ungrammatical for 3rd person (haha) + psych predicate
            print(f"{subject_jp}は{predicate_jp}")

            # Generate evidential form (-sou)
            if predicate_jp.endswith("い"):
                 # Adjective stem + そう
                sou_form = predicate_jp[:-1] + "そう"
                print(f"{subject_jp}は{sou_form}")

if __name__ == "__main__":
    generate_sentences()
\end{verbatim}
\end{CJK}
\caption{Python code to generate example sentences.}
\label{fig:pythoncode}
\end{figure*}

\subsection{Predicates Used in Section~\ref{sec:psychrestrictionperplexityexps} Psych Predicate Experiments}

\begin{tabular}{>{\raggedright}p{1.5cm} p{5cm}}
yasashi & nice \\
kawai & cute \\
oso & late \\
haya & fast \\
chika & close \\
waka & young \\
omoshiro & interesting \\
segataka & tall \\
urusa & loud \\
tsuyo & strong \\
ita & feel.pain \\
samu & feel.cold \\
tsura & feel.difficult \\
kokoroboso & feel.helpless \\
atsu & feel.hot \\
kanashi & feel.sad \\
sabishi & feel.lonely \\
ureshi & feel.happy \\
hazukashi & feel.embarrassed \\
kurushi & feel.strenuous \\
\end{tabular}








\end{document}